\documentclass{amia}
\usepackage{etoolbox}
\usepackage{float}
\usepackage{amsmath}
\usepackage{booktabs}
\usepackage{multirow}
\usepackage{enumitem}

\begin{document}

\title{Predicting Early Functional Decline from Longitudinal Laboratory and Vital Sign Trajectories: A Large-Scale Study Using the All of Us Research Program}

\author{
Rashmita Kudamala, BPT, MS$^{1}$,
Aravind V. Kuruvikkattil, BAMS, MS$^{1}$,
Lalitha Pranathi Pulavarthy, BDS, MS$^{1}$,
Saptarshi Purkayastha, PhD$^{1}$
}

\institutes{
$^{1}$ Dept. of Biomedical Engineering and Informatics, Luddy School of Informatics, Computing, and Engineering,
Indiana University Indianapolis, IN, USA
}

\maketitle

\section*{Abstract}

\textit{%
Functional decline in older adults is typically recognized only after falls or observable gait impairment, closing the window for prevention. We investigated whether temporal trajectories of routine biomarkers --- already recorded but rarely analyzed longitudinally --- can identify patients in the pre-clinical phase of mobility decline. Using the All of Us Research Program (N\,=\,297,861; 11.1\% cases), we derived trajectory features (slope, variability, delta, mean) for twelve biomarkers over a three-year pre-index window. LightGBM models incorporating trajectories significantly outperformed static laboratory summaries (AUROC 0.797 vs.\ 0.755; DeLong $p < 0.001$; AUPRC 0.380 vs.\ 0.304). A 1:1 age- and sex-matched analysis confirmed an independent trajectory signal (AUROC 0.727 vs.\ demographics-only 0.680). A horizon analysis demonstrated sustained prediction 3--12 months before decline onset (AUROC 0.768--0.740). Because the model uses only measurements already ordered in routine care, it supports passive, zero-burden EHR integration for early detection of pre-clinical functional decline.%
}

\section*{Introduction}

Consider a 67-year-old patient seen in primary care. Her hemoglobin has declined from 13.2 to 11.8\,g/dL over three years - each value individually unremarkable, never triggering a clinical alert. Her albumin has drifted from 4.1 to 3.6\,g/dL. Her diastolic blood pressure variability has gradually increased. At each visit, these measurements were interpreted in isolation, assessed for acute abnormalities, and set aside. Two years later, she presents after a fall. The warning signals had been accumulating in the EHR the entire time.

Functional decline - encompassing gait impairment, reduced mobility, and fall risk - affects more than one-third of adults over 65 and contributes to approximately three million emergency visits and \$50 billion in annual U.S.\ healthcare expenditures.\cite{cdc_falls} Despite this burden, functional decline lacks an automated early warning system embedded in routine clinical workflows.\cite{studenski2011gait} It is typically recognized only after observable impairment or a fall event, when the pre-clinical prevention window has closed. Existing frailty tools, such as the Fried phenotype~\cite{fried2001frailty}, the Short Physical Performance Battery~\cite{guralnik1994short}, and the Timed Up-and-Go, require structured functional assessments, manual scoring, or dedicated patient interaction not consistently embedded in primary care workflows. Even when administered, they capture a single timepoint rather than the trajectory of physiologic change.

Multiple pathways underlying mobility decline leave measurable traces in routine laboratory data: anemia-related fatigue, vitamin~D deficiency affecting neuromuscular function, hyponatremia impairing balance, glycemic dysregulation driving neuropathy, and albumin-mediated sarcopenia. \cite{menant2012vitamin,renneboog2006mild} Yet, clinical interpretation focuses on whether individual values exceed reference thresholds, not on how they evolve longitudinally. The central insight motivating this work is that the \textit{trajectory} of a biomarker may carry more predictive information than any single value. A hemoglobin of 11.5\,g/dL that has been stable for three years represents a fundamentally different risk than one that declined from 13.5\,g/dL over the same period. Yet, both appear identical at a single timepoint.

Recent work in EHR-based temporal modeling supports this hypothesis across other clinical domains. Transformer-based architectures modeling sequential diagnosis codes have improved the prediction of ICU mortality, heart failure readmission, and dementia onset by explicitly capturing temporal dependencies.\cite{carrasco2023prediction} In fall prediction specifically, multi-horizon explainable AI models using EHR features achieve AUROCs of 0.70--0.87, with SHAP-based temporal analysis revealing that acute factors drive short-term risk, while long-term risk reflects chronic physiologic deterioration.\cite{pronk2024explainable} However, these models rely on broad feature sets: prior falls, medications, diagnoses, comorbidity scores, and functional assessments, rather than isolating the specific contribution of biomarker trajectories. Meanwhile, statistical joint modeling frameworks for longitudinal biomarker data and survival outcomes face computational limitations when the number of markers exceeds five.\cite{li2021pseudo} A critical gap therefore exists: no large-scale study has systematically evaluated whether temporal trajectory features derived exclusively from routine laboratory measurements, without diagnosis codes, medications, or functional assessments, can predict pre-clinical functional decline.

The All of Us Research Program provides an unprecedented opportunity to address this gap at a national scale, with deliberate oversampling of populations historically underrepresented in biomedical research.\cite{allofus2019} Our contributions are: (1) a pseudo-index cohort design preventing immortal time bias;\cite{suissa2008immortal} (2) trajectory features including slope, variability, and directional change for twelve routine biomarkers; (3) a four-algorithm comparative evaluation using out-of-fold predictions with DeLong significance testing;\cite{delong1988comparing} (4) a 1:1 age- and sex-matched sensitivity analysis isolating independent trajectory signal from demographic confounding; (5) a count-feature removal analysis disentangling physiologic signal from healthcare utilization artifacts - an instance of ``informative missingness" where measurement frequency itself correlates with disease severity;\cite{yang2026joint} (6) SHAP interpretability computed exclusively on held-out test data; and (7) a prediction horizon analysis demonstrating sustained discrimination 3-12 months before the index date (AUROC 0.768--0.740), quantifying the actionable early-detection window available for pre-clinical intervention.

\section*{Methods}

\subsection*{Dataset and Study Population}
We used the All of Us Research Program Controlled Tier Dataset version~8, a de-identified longitudinal EHR repository harmonized to the OMOP Common Data Model. Eligible participants were aged 50-125 years with available EHR data and at least one recorded laboratory measurement. Deceased participants were excluded from all analyses by removing individuals with a recorded entry in the OMOP \texttt{death} table prior to the analysis date. All analyses were conducted within the All of Us Researcher Workbench. The SQL queries used for cohort extraction and biomarker retrieval from OMOP Common Data Model databases are publicly available at: \url{https://github.com/rashmitakudamala/ehr-trajectory-functional-decline}.

\noindent\textbf{Outcome definition.}~Mobility decline cases were identified from the OMOP \texttt{condition\_occurrence} table using ICD-10-CM source concept identifiers: difficulty in walking (R26.2), unspecified gait and mobility abnormalities (R26.9), unsteadiness on feet (R26.81), other reduced mobility (Z74.09), generalized muscle weakness (M62.81), and unspecified fall (W19). These codes were selected based on established EHR-based functional decline operationalizations \cite{stineman2014function, quan2005coding}. The earliest qualifying diagnosis defined each case's index date.

\noindent\textbf{Control cohort and pseudo-index assignment.}~Controls were participants with laboratory data and no qualifying mobility decline diagnosis. A pseudo-index date was assigned by sampling uniformly at random within each control's laboratory observation window (first to last measurement). This design preserves natural measurement history while preventing immortal time bias\cite{suissa2008immortal}, a recognized source of inflated effect estimates in observational studies where controls are implicitly guaranteed survival to a fixed endpoint. The full pseudo-index cohort comprised 322,409 individuals (35,049 cases, 10.9\%).

\subsection*{Biomarker Selection and Observation Windows}

Twelve biomarkers with established pathophysiologic links to functional decline were selected \textit{a priori}: hemoglobin (anemia-fatigue pathway), albumin (frailty-sarcopenia indicator\cite{zhang2024albumin}), sodium (hyponatremia-gait instability\cite{renneboog2006mild}), potassium (neuromuscular excitability\cite{tachi2014hypokalemia}), creatinine (sarcopenia proxy), HbA1c (peripheral neuropathy), 25-hydroxyvitamin~D (neuromuscular function), erythrocyte distribution width (RDW; systemic inflammation and nutritional status marker\cite{liang2022rdw}), C-reactive protein (CRP; chronic low-grade inflammation driving muscle fiber degradation and mobility disability\cite{verghese2012crp}), BMI (body composition trajectory), systolic blood pressure, and diastolic blood pressure (orthostatic dysregulation\cite{mol2019orthostatic}). Measurements were extracted from the OMOP \texttt{measurement} domain using standard concept identifiers and filtered to clinically plausible ranges with required units: BMI 10-80\,kg/m$^2$; systolic BP 70-250\,mmHg; diastolic BP 40-150\,mmHg; sodium 110-170\,mmol/L; potassium 2-7\,mmol/L; creatinine 0.2-15\,mg/dL; hemoglobin 50-200\,g/L; albumin 10-60\,g/L; HbA1c 3-18\%; 25-hydroxyvitamin~D 5-150\,ng/mL; RDW 10-30\%; and CRP 0-300\,mg/L. Equivalent concept name variants (e.g., hemoglobin A1c measured by HPLC vs.\ immunoassay; 25-hydroxyvitamin~D3 vs.\ combined D2+D3; RDW by automated count vs.\ ratio; high-sensitivity CRP vs.\ standard CRP) were harmonized to single canonical labels prior to feature aggregation. All measurements on or after the index date were excluded to prevent data leakage. The primary analysis used a three-year pre-index window (N\,=\,297,861; 33,088 cases, 11.1\%).

\subsection*{Feature Engineering}

Two representations were constructed. \textit{Baseline features} used each biomarker's mean value within the window. \textit{Trajectory features} captured five per-biomarker descriptors: mean, standard deviation (intra-individual variability), delta (last minus first, directional change), count (measurement frequency), and linear slope (rate of change per day). Slope was computed via the ordinary least-squares estimator applied vectorially:

\[
\hat{\beta} = \frac{n\!\sum x_i y_i - \!\sum x_i \!\sum y_i}{n\!\sum x_i^2 - \!\bigl(\!\sum x_i\bigr)^{\!2}}
\]

where $x_i$ denotes days before index and $y_i$ the biomarker value. This yields 80 trajectory features per participant across all concept name variants. Age at index and one-hot encoded gender were included in the combined models. Missing values were imputed with the median of each biomarker computed from the training partition of each cross-validation fold, ensuring held-out data never informed the imputed values and preventing data leakage. Notably, inclusion of measurement count as a feature explicitly models \textit{informative observation} - the phenomenon where biomarker measurement frequency in EHR data is driven by patient health status rather than occurring at random intervals.\cite{yang2026joint}

\subsection*{Model Development and Evaluation}

Four algorithms were evaluated: (1)~$\ell_2$-penalized logistic regression (SAGA solver, \texttt{max\_iter\,=\,1000}, \\ \texttt{class\_weight\,=\,`balanced'}); (2)~random forest (200 estimators, balanced class weights, \texttt{n\_jobs\,=\,$-$1}); (3)~XGBoost (300 estimators, learning rate 0.05, \texttt{eval\_metric\,=\,`auc'});\cite{chen2016xgboost} and (4)~LightGBM (300 estimators, learning rate 0.05).\cite{ke2017lightgbm} Gradient boosting models applied cost-sensitive learning via \texttt{scale\_pos\_weight} set to the per-fold control-to-case ratio; logistic regression features were standardized (\texttt{StandardScaler}) prior to fitting. All models were trained with 5-fold stratified cross-validation; out-of-fold (OOF) predictions across all folds were aggregated for unbiased evaluation. Primary discrimination was assessed by AUROC on OOF predictions with bootstrap 95\% CIs (2,000 resampling iterations). Pairwise model comparisons used the DeLong test for correlated ROC curves\cite{delong1988comparing} applied to paired OOF predictions. Given the 11.1\% case prevalence, the area under the precision-recall curve (AUPRC, equivalent to average precision) was additionally computed as a metric more sensitive to positive-class detection under class imbalance. Calibration was evaluated using Brier scores and reliability curves; isotonic regression calibration was applied via \texttt{CalibratedClassifierCV} ($cv$\,=\,5) on a held-out 20\% stratified test split. Category-based NRI was computed at risk thresholds of 0.10, 0.15, 0.20, 0.25, and 0.30.

\subsection*{Sensitivity Analyses}

\paragraph{Age/sex-matched cohort.} To isolate trajectory signal from demographic confounding, 1:1 greedy matching on gender (exact) and age ($\pm$2 years) was performed without replacement. Matched controls received their paired case's index date.

\paragraph{Utilization feature removal.} All measurement count features were removed, and the DeLong test compared full versus count-removed models - directly testing whether the model exploits informative observation patterns or genuine physiologic dynamics.

\paragraph{Missingness analysis.} Missingness rates per biomarker were computed, stratified by case/control status, to quantify differential measurement availability.

\noindent\textbf{Model interpretability.}~SHAP TreeExplainer values\cite{lundberg2017unified} were computed on a held-out 20\% test set ($n$\,=\,5,000) to prevent training-data leakage - a methodological concern increasingly recognized in predictive modeling.\cite{deng2025auc} Mean absolute SHAP values ranked features globally; metric-level importance was summarized by averaging $|\text{SHAP}|$ across trajectory descriptor types.

\noindent\textbf{Ethics and data governance.}~This study used data from the All of Us Research Program's Controlled Tier Dataset v8, available to authorized users on the Researcher Workbench. The All of Us Research Program operates under a central IRB protocol; no additional ethics review is required for analyses conducted within the Researcher Workbench.

\section*{Results}

\subsection*{Study Population and Missingness}

\begin{figure}[H]
\centering
\includegraphics[width=0.70\columnwidth]{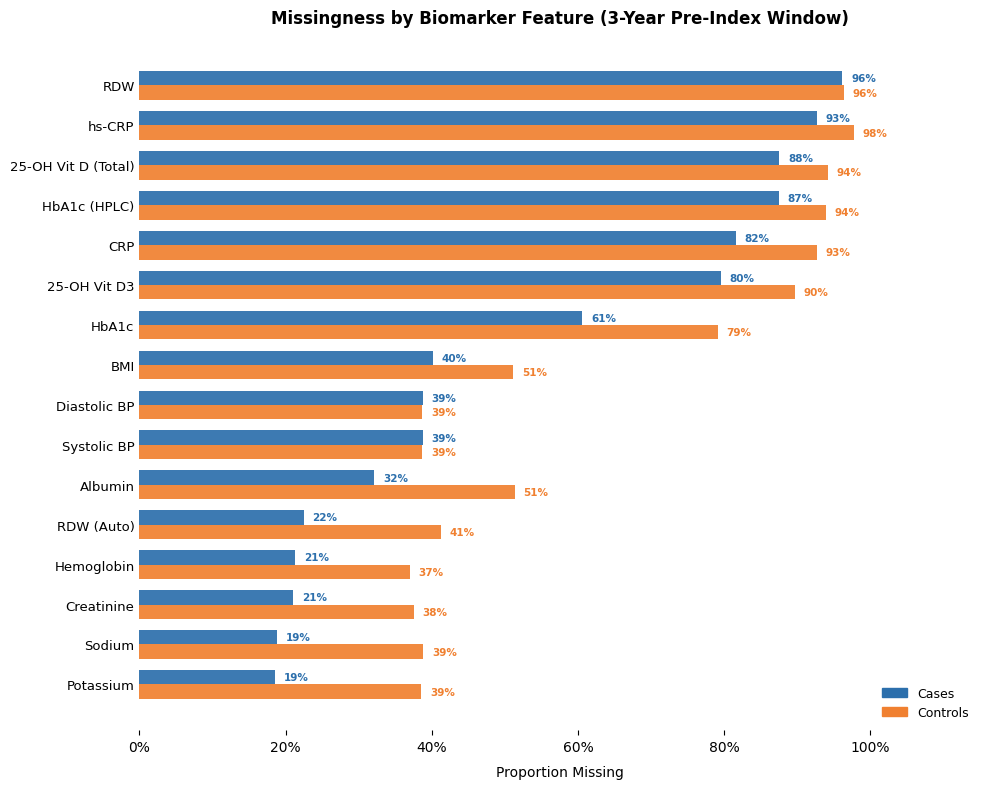}
\caption{Differential missingness by lab feature, stratified by outcome (3-year window). Cases show consistently lower missingness, consistent with informative observation in EHR data - patients developing functional impairment receive more frequent monitoring.}
\label{fig:missingness}
\end{figure}

The primary modeling cohort contained 297,861 participants (33,088 cases, 11.1\%). Mean age at index was 50.6 years (SD\,=\,16.8). Gender distribution was 61.7\% female and 36.5\% male. Cases showed systematically lower missingness than controls across all biomarkers (Figure~\ref{fig:missingness}), with the largest differential for albumin (32.1\% vs.\ 51.4\%) and hemoglobin (21.3\% vs.\ 37.0\%). This pattern exemplifies informative observation: patients developing functional impairment receive more frequent monitoring, making measurement frequency a confounder rather than a neutral feature.\cite{yang2026joint} CRP (82\%--98\% missing across variants) and 25-hydroxyvitamin~D (80\%--88\%) exhibited the greatest overall sparsity, limiting their trajectory signal. The three highest-ranked physiologic predictors, BMI mean (0.141), diastolic blood pressure slope (0.104), and RDW mean (0.094), each had substantial missingness (Figure~\ref{fig:missingness}), meaning their SHAP values likely underestimate their true predictive contributions in more completely monitored populations.

\subsection*{Trajectory Features Improve Discrimination}

\begin{figure}[H]
\centering
\includegraphics[width=0.50\columnwidth]{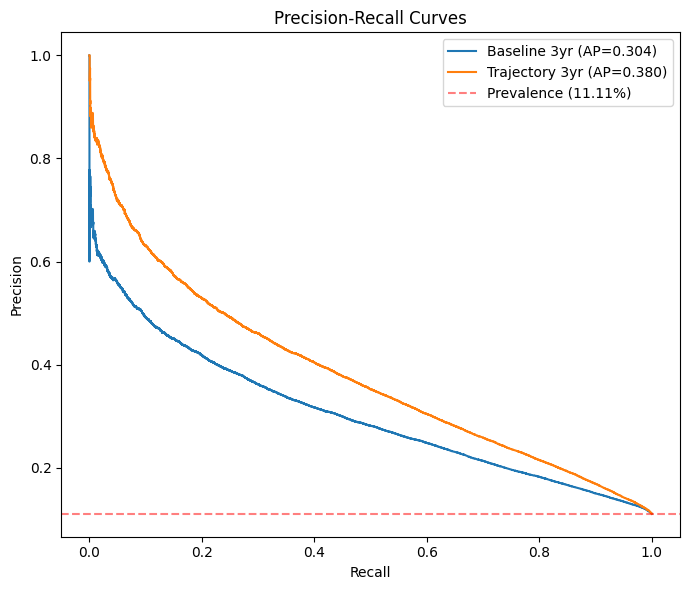}
\caption{Precision--recall curves. The trajectory model (AP\,=\,0.380) achieves 25\% relative improvement over baseline (AP\,=\,0.304). Dashed line: 11.1\% prevalence.}
\label{fig:pr}
\end{figure}

Table~\ref{tab:performance} presents model performance. A stepwise improvement emerged as temporal information was added. Demographics alone achieved AUROC 0.680 (95\% CI: 0.677--0.683). Baseline laboratory means improved to 0.755 (0.752--0.757). Trajectory features reached 0.797 (0.794--0.800), a significant gain over baseline ($\Delta$AUC\,=\,0.042; DeLong $p < 0.001$). Trajectory AUPRC was 0.380 versus 0.304 for baseline - a 25\% relative improvement and 3.4$\times$ the no-skill rate of 0.111 (Figure~\ref{fig:pr}). The larger relative AUPRC gain (+0.076) than AUROC gain (+0.042) indicates that trajectories disproportionately improve positive-class detection at high-precision operating points - critical for alert systems where false positives drive clinician fatigue.\cite{deng2025auc}

\begin{table}[H]
\caption{Model performance (3-year window, 5-fold CV, OOF; N\,=\,297,861; 11.1\% cases). Bootstrap 95\% CI and AUPRC reported for LightGBM. LR\,=\,logistic regression; RF\,=\,random forest; XGB\,=\,XGBoost; LGB\,=\,LightGBM.}
\label{tab:performance}
\begin{center}
\footnotesize
\begin{tabular}{@{}llccc@{}}
\toprule
\textbf{Feature Set} & \textbf{Model} & \textbf{AUC} & \textbf{95\% CI} & \textbf{AUPRC} \\
\midrule
Demographics only & LGB & 0.680 & [0.677--0.683] & 0.204 \\
\midrule
\multirow{4}{*}{Baseline labs (3yr)} & LR  & 0.640 & -- & -- \\
 & RF  & 0.714 & -- & -- \\
 & XGB & 0.750 & -- & -- \\
 & \textbf{LGB} & \textbf{0.755} & \textbf{[0.752--0.757]} & \textbf{0.304} \\
\midrule
\multirow{4}{*}{Trajectory (3yr)} & LR  & 0.726 & -- & -- \\
 & RF  & 0.775 & -- & -- \\
 & XGB & 0.795 & -- & -- \\
 & \textbf{LGB} & \textbf{0.797} & \textbf{[0.794--0.800]} & \textbf{0.380} \\
\midrule
Trajectory (matched) & LGB & 0.727 & [0.723--0.731] & -- \\
\bottomrule
\end{tabular}
\end{center}
\end{table}

\noindent\textbf{Matched cohort: isolating trajectory signal.}~Of 35,049 cases, the vast majority were successfully matched ($<$20 unmatched). The matched trajectory model achieved AUROC 0.727 (0.723--0.731). This result enables a three-way signal decomposition: of the total 0.797 AUROC, 0.180 units above chance (0.5) are attributable to demographics (0.680 $-$ 0.5), 0.047 to independent laboratory trajectory signal (0.727 $-$ 0.680), and 0.070 to demographic confounding removed by matching (0.797 $-$ 0.727). The retained 0.727 in the matched cohort directly confirms that biomarker trajectories capture genuine physiologic information not explained by age and sex.

\noindent\textbf{Prediction horizon analysis.}~To evaluate the temporal lead time available for early detection, we assessed model discrimination when laboratory measurements within 3, 6, 9, and 12 months of the index date were excluded. This design simulates prospective screening where predictions must be made before clinical recognition of mobility decline.

Trajectory-based models maintained substantial discrimination across all horizons. AUROC declined gradually from 0.768 at the 3-month horizon to 0.740 at the 12-month horizon, demonstrating that biomarker trajectories contain predictive signal up to one year before diagnosis (Table~\ref{tab:horizon}). Corresponding AUPRC values remained stable (0.361–0.353), indicating consistent positive-class detection despite increasing temporal distance from the outcome.

These results quantify a clinically meaningful pre-clinical window during which physiologic deterioration is detectable in routine laboratory data before functional decline becomes clinically documented.

\begin{table}[H]
\caption{Prediction horizon performance using trajectory features. Measurements within the specified horizon prior to the index date were excluded.}
\label{tab:horizon}
\centering
\small
\begin{tabular}{ccc}
\toprule
\textbf{Prediction Horizon} & \textbf{AUROC} & \textbf{AUPRC} \\
\midrule
3 months  & 0.768 & 0.361 \\
6 months  & 0.756 & 0.358 \\
9 months  & 0.748 & 0.355 \\
12 months & 0.740 & 0.353 \\
\bottomrule
\end{tabular}
\end{table}

\begin{figure}[!bhtp]
\centering
\includegraphics[width=0.99\columnwidth]{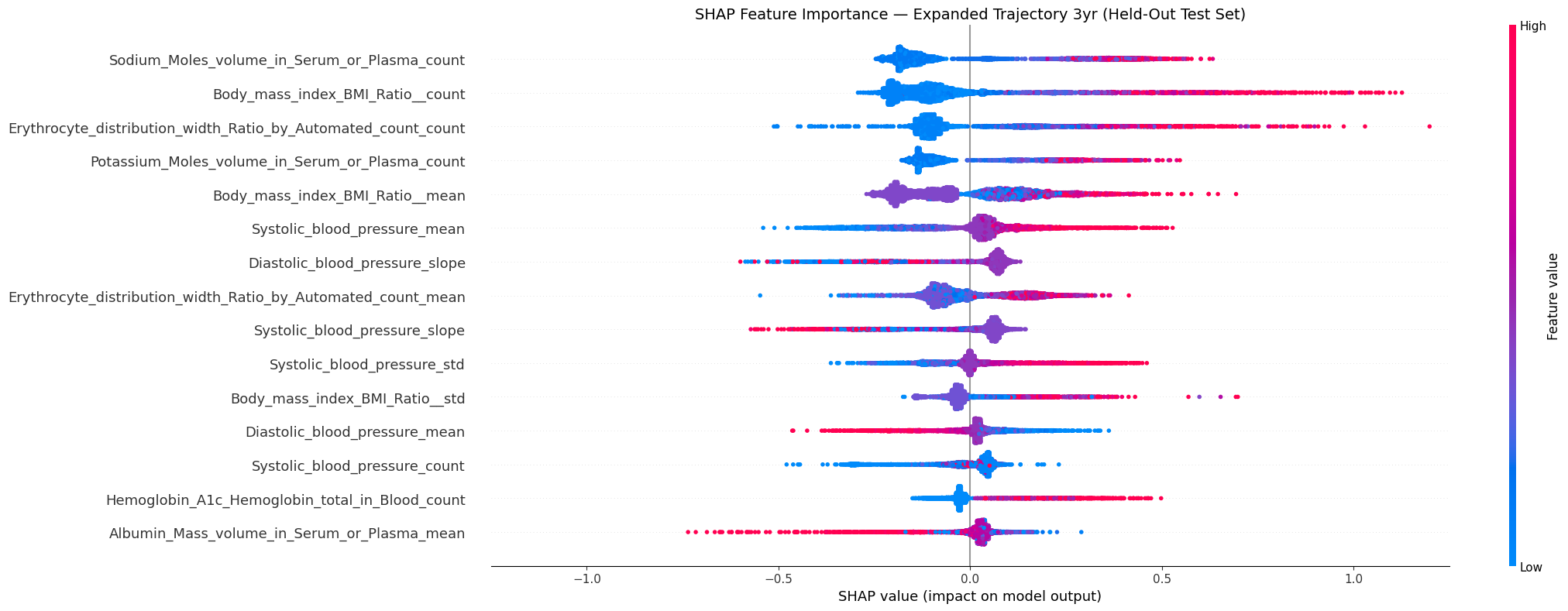}
\caption{SHAP beeswarm plot (held-out test set, $n$\,=\,5,000). Red: high feature values. BMI mean, blood pressure slopes, RDW mean, and systolic BP mean appear among the top physiologic predictors, with trajectory-specific features (slope, std) complementing static means.}
\label{fig:shap}
\end{figure}

\noindent\textbf{Feature importance.}~By mean $|\text{SHAP}|$ on 5,000 held-out observations, the top physiologic trajectory features were BMI mean (0.141), systolic blood pressure mean (0.113), diastolic blood pressure slope (0.104), RDW mean (0.094), and systolic blood pressure slope (0.088) (Figure~\ref{fig:shap}). Count features ranked highest overall (sodium count: 0.198; BMI count: 0.189; RDW count: 0.161; potassium count: 0.142), consistent with informative observation. Aggregating by metric type: count (mean $|\text{SHAP}|$\,=\,0.061), mean (0.047), slope (0.025), std (0.017), delta (0.009).

The prominence of diastolic blood pressure slope as the leading trajectory-specific feature (i.e., highest-ranked among slope, std, and delta descriptors) aligns with growing evidence that blood pressure dysregulation - particularly orthostatic hypotension, which affects approximately 20\% of community-dwelling older adults\cite{mol2019orthostatic} - is a major intrinsic risk factor for falls.\cite{pronk2024cardiovascular} RDW mean (0.094) ranked fourth overall among non-count features, consistent with recent evidence linking erythrocyte distribution width to frailty and all-cause mortality independently of hemoglobin.\cite{liang2022rdw} BMI mean was the single most important non-count feature, reflecting the central role of body composition in sarcopenia-mediated mobility decline.

\noindent\textbf{Sensitivity analyses and calibration.}~Removing count features yielded AUROC 0.789 (0.786--0.791), a reduction of 0.009 (DeLong $p < 0.001$). While statistically significant due to large sample size, the magnitude confirms that the model primarily exploits physiologic dynamics. This is clinically important: it means the model performs comparably in settings with variable healthcare utilization, including safety-net hospitals and rural practices where measurement frequency does not reliably reflect disease severity.

\begin{figure}[!bhtp]
\centering
\includegraphics[width=0.50\columnwidth]{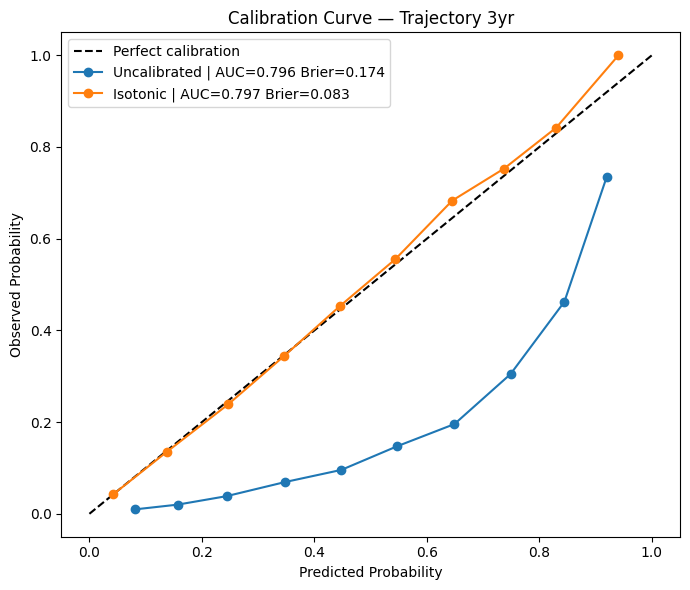}
\caption{Calibration curves. Isotonic regression substantially improves reliability (Brier 0.083 vs.\ 0.174) with AUC preserved.}
\label{fig:calibration}
\end{figure}

Isotonic calibration reduced Brier score from 0.174 to 0.083 (52.3\% improvement) without degrading discrimination (AUROC 0.797 vs.\ 0.796; Figure~\ref{fig:calibration}). At risk thresholds of 0.20 and 0.25, the trajectory model achieved NRI of 0.077 and 0.124 over baseline, with gains concentrated in improved specificity among non-events (NRI$_\text{nonevents}$\,=\,0.091 and 0.153), indicating reduced unnecessary risk flagging.

\section*{Discussion}

We demonstrate that longitudinal biomarker trajectories from twelve routine measurements significantly improve prediction of functional mobility decline compared with static summaries (AUROC 0.797 vs.\ 0.755; AUPRC 0.380 vs.\ 0.304; $p < 0.001$). A matched analysis confirmed an independent trajectory signal (AUROC 0.727), and isotonic calibration produced well-calibrated probabilities (Brier 0.083). These results support the hypothesis that functional decline unfolds through gradual physiologic instability already being recorded, and currently being ignored, in routine EHR data. The horizon analysis further demonstrates that these signals are detectable up to 12 months before diagnosis, suggesting a substantial window for preventive intervention.

\paragraph{Trajectory features as passive digital biomarkers.} Our approach is distinguished from existing fall prediction models by its exclusive reliance on structured laboratory data. Recent multi-horizon EHR models for fall risk achieve AUROCs of 0.70--0.87, but require diagnosis histories, medication lists, comorbidity scores, and prior fall documentation.\cite{pronk2024explainable,yousufuddin2025inpatient} Our model achieves AUROC 0.797 using only twelve laboratory biomarkers - no diagnosis codes, medications, functional assessments, or free-text processing. This constraint is a deliberate design choice: it produces a model that functions as a \textit{passive digital biomarker}, requiring no new data collection, patient interaction, or clinical workflow modification. Transformer-based approaches that model full longitudinal EHR sequences\cite{carrasco2023prediction} capture richer temporal dependencies but require sequence-level architectures and training infrastructure that may limit deployment in resource-constrained settings. Our feature-engineered approach offers a pragmatic middle ground: it captures the key temporal dynamics (slope, variability, directional change) that distinguish declining patients while remaining compatible with standard tabular classifiers deployable in any EHR environment.

\paragraph{Mechanistic interpretation.} Among non-count features, BMI mean was the highest-ranked predictor \\(mean $|\text{SHAP}|$\,=\,0.141), consistent with the central role of body composition in sarcopenia-mediated mobility decline: both obesity (reduced mobility, joint loading) and low BMI (muscle wasting) represent distinct but convergent pathways to functional impairment.\cite{studenski2011gait} Diastolic blood pressure slope, the leading trajectory-specific feature (0.104), connects to a well-characterized autonomic pathway. Orthostatic hypotension affects $\sim$20\% of community-dwelling older adults and is significantly associated with fall risk (meta-analytic OR\,=\,1.73; 95\% CI 1.50--1.99).\cite{mol2019orthostatic} A declining diastolic trajectory captured from routine seated recordings may reflect progressive autonomic dysregulation before orthostatic symptoms become clinically apparent.\cite{pronk2024cardiovascular} RDW mean (0.094) is a notable finding: elevated RDW reflects erythropoietic instability driven by chronic inflammation, iron deficiency, and oxidative stress, and has been independently associated with frailty, all-cause mortality, and functional decline in older adults.\cite{liang2022rdw} Unlike hemoglobin, which captures oxygen-carrying capacity, RDW captures the heterogeneity of the red cell population, a marker of systemic physiologic dysregulation that may precede overt anemia. The co-presence of trajectory-specific features (slope, std) and static means in the SHAP top-15 directly confirms that the same biomarker level carries different risk depending on its trajectory.

\paragraph{Why AUPRC improvement matters more than AUROC improvement.}
The relative gain in AUPRC (+0.076, 25\%) substantially exceeds the relative gain in AUROC (+0.042, 6\%), and this asymmetry is clinically informative. With 11\% case prevalence, AUROC can remain high even when positive-class detection is poor, because it is dominated by the large number of true negatives among controls. AUPRC specifically rewards improved precision at any given recall level - meaning that the trajectory model correctly identifies more true mobility decline cases while generating fewer false positive alerts per true positive detected. In EHR-integrated alert systems, where clinician alert fatigue is a primary barrier to adoption,\cite{deng2025auc} this distinction directly predicts greater real-world utility. The trajectory model's AUPRC of 0.380 - approximately 3.4$\times$ the no-skill baseline of 0.111 - suggests that at moderate recall levels, roughly one in three flagged patients would be a true positive, a ratio far more compatible with clinical workflow integration than the baseline model's one-in-four rate.

\paragraph{Informative observation as confounder and signal.} A distinctive finding is the high SHAP importance of measurement count features (sodium count: 0.198; BMI count: 0.189; RDW count: 0.161), which outranked all physiologic trajectory features. This reflects the EHR-specific phenomenon of \textit{informative observation}: measurement frequency in EHR data is not random but driven by patient health status, creating a two-stage missingness mechanism where both the visiting process and the biomarker observation process correlate with the outcome.\cite{yang2026joint} Our sensitivity analysis removing count features (AUROC 0.789, $\Delta = -0.009$) demonstrates that while the model can exploit this signal, it does not depend on it. This finding has direct implications for health equity: it means the model retains strong performance even when applied to patients with infrequent healthcare contact, a critical requirement for deployment in safety-net settings, federally qualified health centers, and rural practices serving populations with the highest functional decline burden.

\paragraph{Comparison with prior work.} Prior ML models for functional impairment report AUROCs of 0.87--0.92 using feature sets incorporating falls history, hospitalization data, comorbidity problem lists, and social determinants.\cite{laberge2023ml,feng2023physical} Our model achieves its discrimination using only structured biomarkers, without free-text processing or specialist assessments. The All of Us cohort's deliberate oversampling of Black or African American (15.5\%), Hispanic, and rural participants\cite{allofus2019} addresses a recognized generalizability limitation of prior work, which is largely derived from predominantly White academic medical center populations.

\paragraph{Toward clinical deployment.} The model requires no new tests, no patient interaction, and no additional documentation. It runs as a background process, updating risk scores each time a new laboratory result is filed. At a clinical visit, the physician sees a single alert - \textit{``Elevated functional decline risk - consider physical therapy referral or geriatric assessment''} - while the pre-clinical intervention window remains open. The AUROC 0.789 retained after count feature removal means this workflow performs comparably even for patients seen infrequently. To operationalize alerts, a probability threshold must reflect local referral 
capacity. At 0.20 - where NRI over baseline was 0.077 - roughly one in three flagged patients is a true positive (AUPRC 0.380). At 0.25, where NRI gain was highest (0.124), a higher-specificity alert could trigger comprehensive geriatric assessment. The 12-month detection horizon means alerts arrive well before observable impairment, preserving time for evidence-based fall-prevention programs.

\paragraph{Limitations.} ICD-10-coded outcomes capture documented recognition rather than objective functional performance, introducing potential detection bias. CRP and 25-hydroxyvitamin~D contribute a limited trajectory signal due to selective ordering (82\%--98\% and 80\%--88\% missing, respectively); methods that explicitly model informative missingness mechanisms\cite{yang2026joint} may improve these feature contributions in future work. Median imputation assigns a population-level central value regardless of clinical context; future studies should assess sensitivity to the assumed missingness mechanism and consider multiple imputation approaches. All evaluations use cross-sectional random splits; calendar-time validation would strengthen temporal generalizability evidence. Comorbidity adjustment was not performed. The model may partially capture known disease trajectories rather than purely pre-clinical signals. External validation in independent health systems is required before deployment, particularly given that biomarkers were selected \textit{a priori} based on known pathophysiologic links - validation against cohorts using objective functional endpoints such as gait speed or SPPB would test generalizability beyond administratively-defined decline.

\section*{Conclusion}

Longitudinal trajectory features from twelve routine biomarkers significantly outperform static laboratory summaries for predicting functional mobility decline (AUROC 0.797 vs.\ 0.755; $p < 0.001$; AUPRC 0.380 vs.\ 0.304). A matched sensitivity analysis confirms an independent trajectory signal beyond demographic confounding (AUROC 0.727 vs.\ demographics-only 0.680), and removal of utilization-proxy features preserves 99\% of discrimination (AUROC 0.789). Because the model relies exclusively on measurements already ordered in routine primary care, trajectory-based risk monitoring functions as a passive digital biomarker supporting equitable, zero-burden integration into EHR workflows for early detection of pre-clinical functional decline.

\section*{Acknowledgements}

We gratefully acknowledge All of Us participants for their contributions, without whom this research would not have been possible. We also thank the National Institutes of Health's All of Us Research Program for making available the participant data examined in this study.
\bibliographystyle{unsrt}
\bibliography{amia}

\end{document}